\documentclass[letterpaper, 10 pt, conference]{ieeeconf}  

\IEEEoverridecommandlockouts                              

\usepackage{graphics} 
\usepackage{graphicx}
\usepackage{epsfig} 

\usepackage{tabularx}
\usepackage{soul} 

\usepackage[font=footnotesize]{caption}
\usepackage[font=footnotesize]{subcaption}
\usepackage[utf8]{inputenc}
\usepackage[export]{adjustbox}
\usepackage{wrapfig}
\usepackage{array}
\usepackage{multirow}
\usepackage{booktabs}
\usepackage{float}
\usepackage{cite} 

\usepackage{dsfont}
\usepackage{url}
\usepackage{amsmath}
\usepackage{xcolor}

\newcommand{\platformName}{SAM}

\usepackage{amsmath}
\usepackage{bm}
\newcommand{\bM}{\bm{M}}
\newcommand{\bC}{\bm{C}}
\newcommand{\bg}{\bm{g}}
\newcommand{\bq}{\bm{q}}
\newcommand{\btau}{\bm{\tau}}
\usepackage{amsmath}
\usepackage{xcolor}

\title{\LARGE \bf
Development of \platformName: cable-Suspended Aerial Manipulator* 
}

\author{Yuri S. Sarkisov$^{1,3,\dagger}$, Min Jun Kim$^{1,\dagger}$, Davide Bicego$^2$, \\
	Dzmitry Tsetserukou$^{3}$, Christian Ott$^{1}$, Antonio Franchi$^2$, and Konstantin Kondak$^{1}$ 
	\thanks{*Patent pending.}
	\thanks{The funding of the European Commission to the AEROARMS project under the H2020 Programme (Grant Agreement 644271) is acknowledged.}
	\thanks{$\dagger$ The authors contributed equally to this work.}
	\thanks{$^{1}$The authors are with Institute of Robotics and Mechatronics, German Aerospace Center (DLR), Wessling, Germany.}
	\thanks{$^{2}$The authors are with LAAS-CNRS, Universit\'e de Toulouse, CNRS, Toulouse, France.}
	\thanks{$^{3}$The authors are with Space CREI, Skolkovo Institute of Science and Technology (Skoltech), Moscow, Russia.}
	\thanks{
	{\tt\footnotesize e-mails: iurii.sarkisov@dlr.de, minjun.kim@dlr.de, davide.bicego@laas.fr, d.tsetserukou@skoltech.ru, Christian.Ott@dlr.de, antonio.franchi@laas.fr, Konstantin.Kondak@dlr.de}}%
}

\graphicspath{ {images/} }

\begin{document}

\maketitle
\thispagestyle{empty}
\pagestyle{empty}

\begin{abstract}
High risk of a collision between rotor blades and the obstacles in a complex environment imposes restrictions on the aerial manipulators. To solve this issue, a novel system cable-Suspended Aerial Manipulator (SAM) is presented in this paper. Instead of attaching a robotic manipulator directly to an aerial carrier, it is mounted on an active platform which is suspended on the carrier by means of a cable. As a result, higher safety can be achieved because the aerial carrier can keep a distance from the obstacles. For self-stabilization, the SAM is equipped with two actuation systems: winches and propulsion units. This paper presents an overview of the SAM including the concept behind, hardware realization, control strategy, and the first experimental results.

\end{abstract}



\section{Introduction}
\label{sec:Introduction}

Aerial manipulation is one of the most prospective examples of unmanned aerial vehicle (UAV) contact applications. 
The development of a system combining robotic manipulator capabilities and UAV dexterity is complex and requires in-depth research, analysis, and significant experience. Among a number of applications for the aerial manipulators \cite{ruggiero2018aerial}, it is essential to list the following: the inspection of various structures, e.g., bridges, electric lines, and pipelines \cite{baturone2018aeroarms}, assembly/repair of remotely located constructions~\cite{staub2018towards}, and any operations in hazardous/dangerous conditions for human safety such as decommissioning of damaged nuclear power plants. More examples with detailed description can be found in \cite{korpela2018aerial}.

Two main branches can be distinguished in this area. First one is the use of the specific mechanism (e.g., gripper) for a particular type of aerial interaction with an environment. For example, in \cite{thomas2013avian}, an avian-inspired aerial vehicle capable of grasping and transporting different objects was demonstrated. A UAV equipped with a brush for cleaning of the vertical surfaces was proposed in \cite{albers2010semi}. In \cite{bernard2009generic}, slung-load transportation and deployment by single and multiple UAVs were demonstrated. Mechanisms that enable compliance interaction with the environment were shown in \cite{keemink2012mechanical, bartelds2016compliant}. 

Another important branch is an integration of a robotic arm (or even multiple arms) into the UAV. Thus, in \cite{kim2013aerial, suarez2016lightweight, korpela2012mm}, light-weight 2-3 degrees of freedom (DoF) manipulators were used to perform uncomplicated tasks. Furthermore, in \cite{jimenez2013control, caraballo2017block}, aerial manipulators were exploited to perform assembly tasks. In order to explore further capabilities of the aerial manipulation, a 7-DoF torque-controlled KUKA LWR robot was integrated into a UAV \cite{huber2013first, kim2018stabilizing}. Although 7-DoF robotic arm provides many useful benefits such as redundancy \cite{lippiello2012exploiting, kondak2014aerial} and full task space formulation \cite{kim2018passive}, due to the heavy weight (around 15 $\mathrm{kg}$), the manipulator was mounted on the autonomous middle-scale helicopter system with 3.7-meter diameter rotor blades. However, with such dimensions, approaching to a target object might be hard and not safe in a complex environment. Moreover, dynamic turbulence caused by ground effect near wide surfaces serves as an additional source of danger for the helicopter and makes it even harder to operate.  

\begin{figure}[t]
	\centering
	\includegraphics[width=1\linewidth]{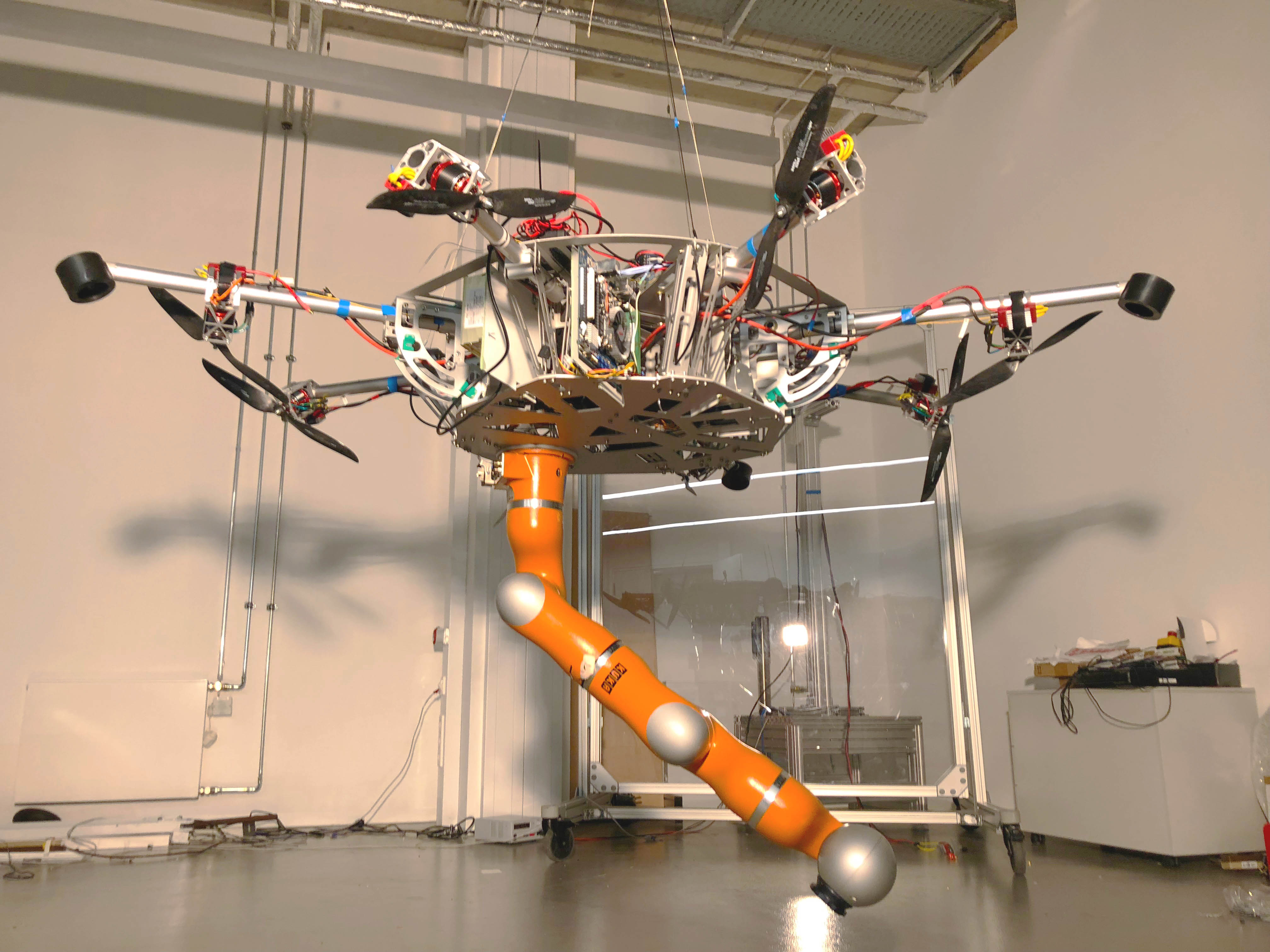}
	\caption{A prototype of the active cable-suspended aerial manipulator \platformName.}
	\label{fig:main}
\end{figure}

To resolve these issues, several prototypes with long reach manipulators have recently been developed \cite{suarez2018flexible, suarez2018light}. Long reach means that the robotic manipulator is attached to the UAV using a long flexible link instead of being mounted directly. It allows to perform the aerial manipulation in a narrow and complex environment while keeping the UAV at a safe distance from the obstacles. However, the use of a flexible link results in underactuation of the system, which complicates the control problem. In \cite{kim2018rope}, a similar system concept was proposed, but with additional moving masses to damp out the oscillations caused by the long reach configuration. However, the use of moving masses does not allow to generate a spatial wrench, which is beneficial in interaction problems. Additionally, moving masses unnecessarily increase the total weight.

In this paper, motivated by the foregoing literature review, we propose a novel active platform called cable-Suspended Aerial Manipulator (\platformName), see Fig. \ref{fig:main}. This system is  equipped with 7-DoF robotic manipulator and has two different means to control its own state: propulsion units and winches. As shown in the schematic diagram in Fig. \ref{fig:cad}, the \platformName ~is hanged on the main aerial carrier (which is an autonomous helicopter\footnote{It is worth mentioning that aerial carrier can be not only a UAV but also a manned aerial vehicle or even a crane. Such variations might be critical for places and applications where the operation of UAVs is restricted.}) utilizing a long cable. At the beginning of the operation, the main aerial carrier should deliver the cable-based platform to the target location. After that, its central role is to hover while the \platformName ~is performing the task independently. The leading feature of this configuration is that the weight of the \platformName ~is supported by the main aerial carrier. Since there is no need to compensate for the gravity, the required amount of the thrust of the propulsion units in the manipulation platform can be reduced. As a result, the diameter of the propeller in the propulsion units can be significantly scaled down. Therefore, the safety of this system is higher. It is worth noting also that the main aerial carrier, which has large rotor blades, can operate far away from the obstacles. Thus, the proposed system alleviates the safety problem.

The main goal of this paper is to give an overview of the \platformName's design, its development process, and control. The paper is organized as follows: Section \ref{sec:The \platformName} gives the main overview of the \platformName ~prototype. Section \ref{sec:control} presents the modeling, preliminary control approach, and the first experimental results of the cable-suspended platform. Finally, Section \ref{sec:end} concludes the paper.

\section{The \platformName}
\label{sec:The \platformName}

As was mentioned above, the \platformName ~is suspended by means of the cable to the main aerial carrier. Therefore, no energy is required for the platform itself to resist gravity. Consequently, the platform can have reasonable dimensions that allow working in a complex environment. The \platformName ~is equipped with winches and propulsion units.
While winches are used to compensate for the slow center of the mass (CoM) displacement change during the manipulation, the propulsion units can be used to reduce dynamic deviations. Such a strategy allows to reduce the amount of energy consumption since no continuous torque is required from the propulsion units.


\subsection{Design of the \platformName}
\label{subsec:dandd}
In this subsection, we will describe the main functional components of the developed platform and present the key features of the \platformName.

\begin{figure}[t]
	\centering
	\includegraphics[width=1\linewidth]{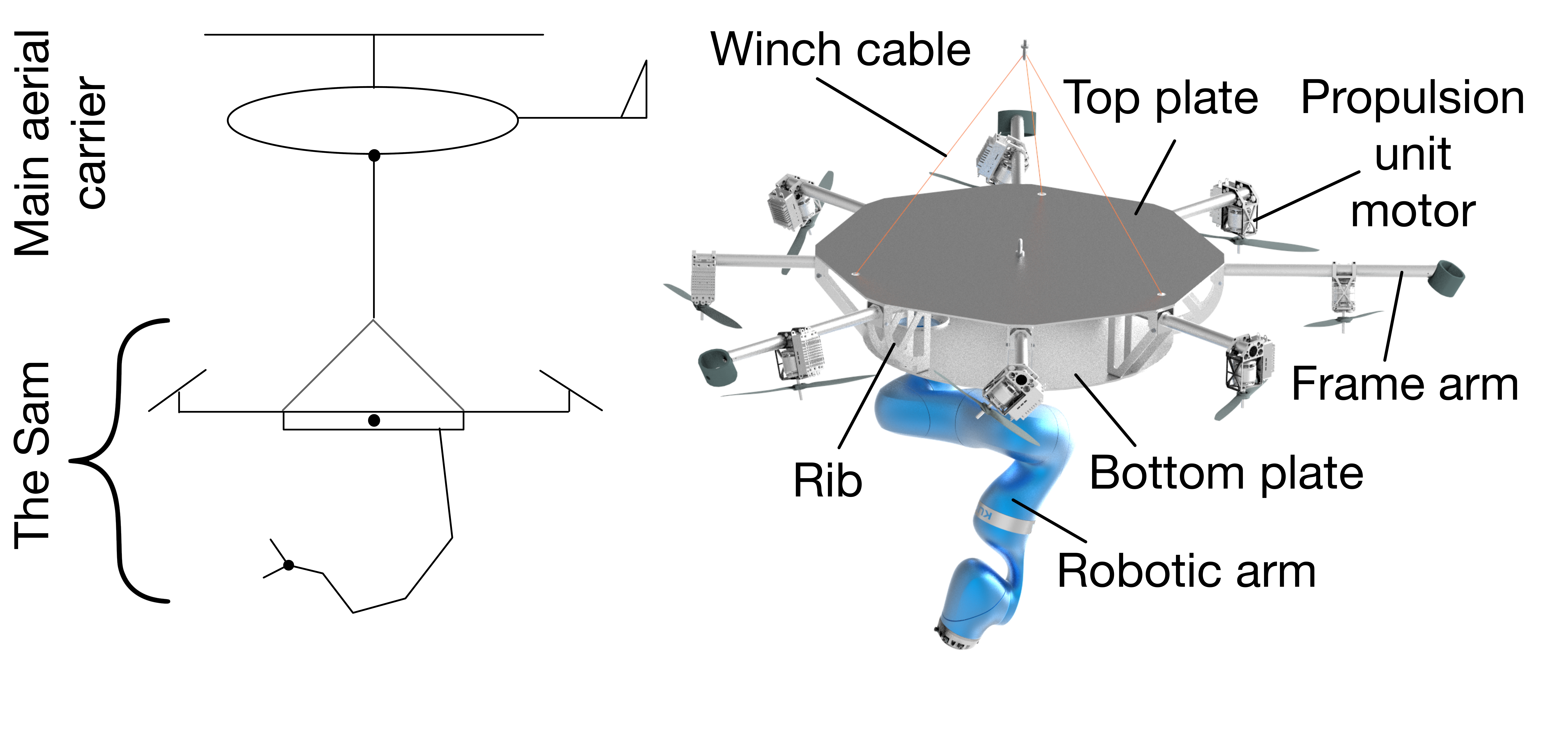}
	\caption{An integration concept (left) and the structure of the \platformName ~(right).}
	\label{fig:cad}
\end{figure}

\subsubsection{Landing gear}
\label{subsubsec:gear}
\begin{figure}[b]
	\centering
	\includegraphics[width=1\linewidth]{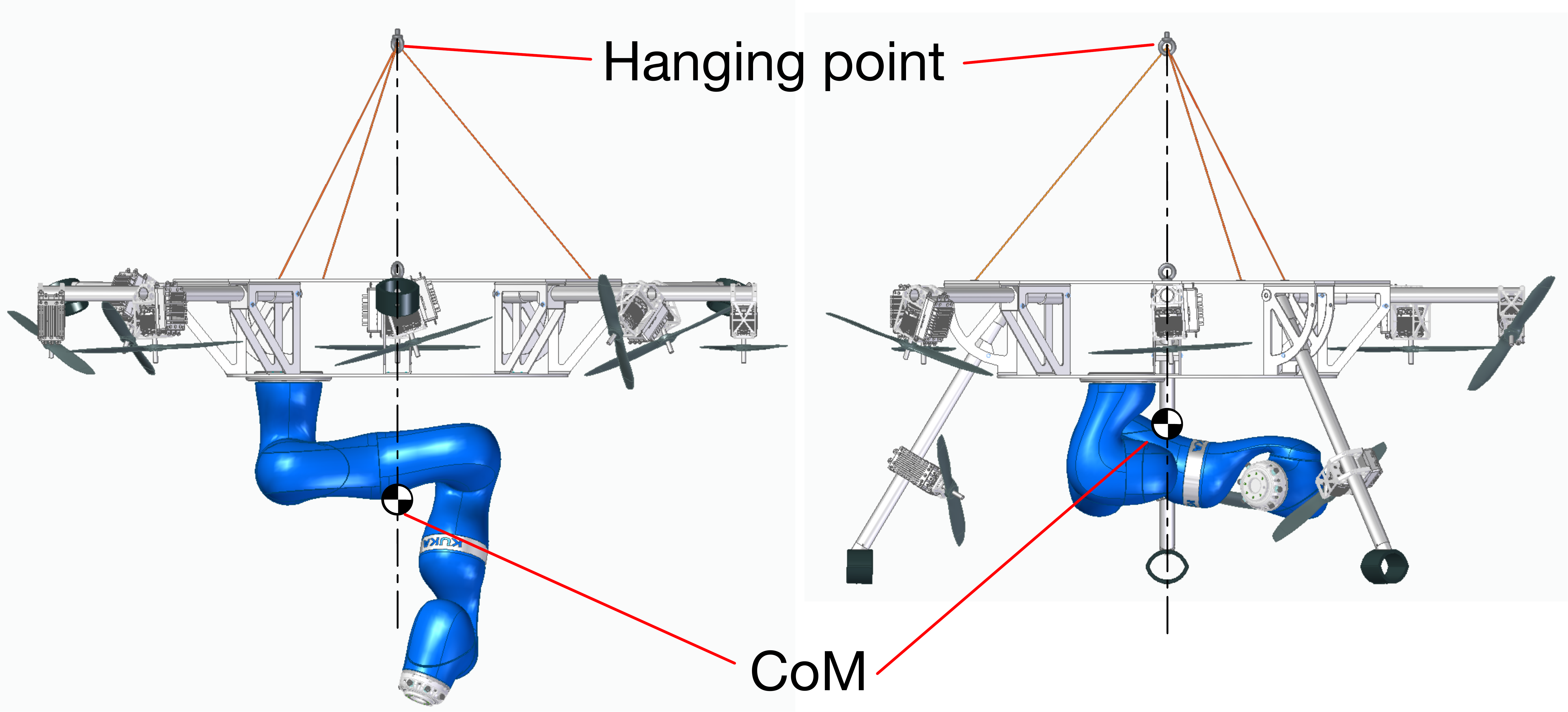}
	\caption{Operation and parking configurations of the \platformName.}
	\label{fig:parking_fin}
\end{figure}

to reduce the total weight and to accommodate large workspace of the manipulator, the traditional landing gear is not installed in the cable-suspended platform. To land, three out of eight frame propeller arms can be folded and converted to the legs of the landing gear, see Fig. \ref{fig:cad} (right) and Fig. \ref{fig:parking_fin}. Thus, the legs have a dual use: landing legs and propeller arms. In order to switch from  the  landing leg to the  propeller  arm,  each  leg  has  to  be  risen  from  the  lower  to  the  upper  position. In  this transition, the leg rotates in the bearing through about 60 degrees. To lift up the leg,  the thrust  force  of  the  propulsion unit is  used. In order to compensate a non-zero wrench caused by mentioned thrust during this procedure, propulsion units of the remaining five propeller arms are used. The movement downwards (transition to the landing leg) is done by the gravity force, damped by the thrust force.

\begin{figure}[t]
	\centering
	\includegraphics[width=1\linewidth]{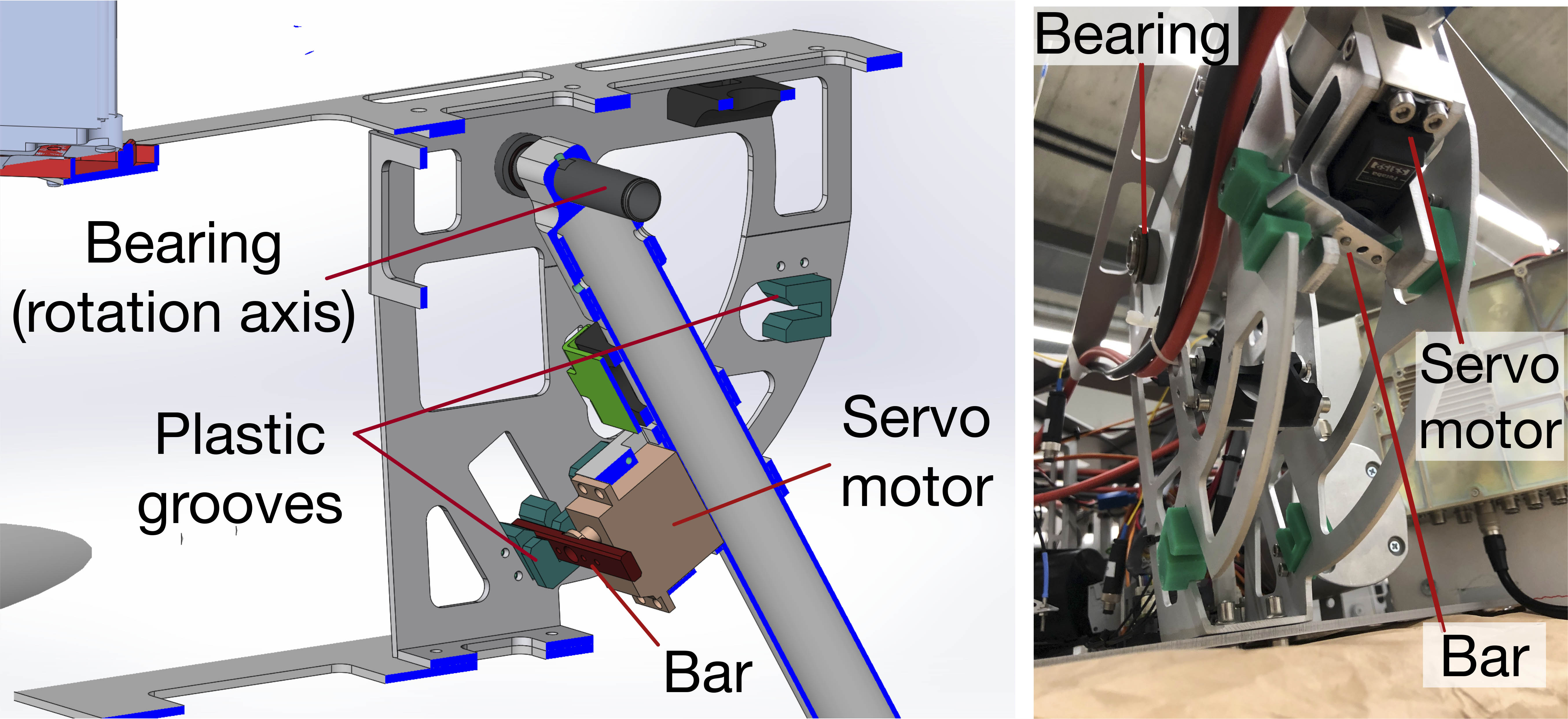}
	\caption{The operating principle of the locker mechanism: CAD model (left) and prototype picture (right).}
	\label{fig:locker}
\end{figure}

To fix the position of the leg/arm in the bottom and top points, the locker mechanism is used, see Fig. \ref{fig:locker}. As it can be seen, the locker mechanism contains two fixed plastic grooves and a servo motor (Futaba S3152). The grooves are manufactured from "S" green plastic material which has low friction, long lifetime, and high wear resistance. In order to lock the position, the servo motor should turn a bar installed at the shaft. 

By virtue of the transformable landing gear, the \platformName ~has a larger workspace for manipulation without any restrictions created by the common landing gear, e.g., skids.

\subsubsection{Robotic manipulator}
\label{subsubsec:arm}
to perform arbitrary manipulation tasks, a 7-DoF KUKA LWR 4 is mounted on the bottom side of the platform. Two main postures of the robotic arm should be defined: operation and parking (Fig. \ref{fig:parking_fin}). During the transportation and landing, the manipulator should be in the parking position. Before initiating any manipulation task such as pick and place or peg-in-hole, manipulator should be placed in the operation configuration as fewer movements are required to perform any typical manipulation task from this position.

Additionally, it is worth mentioning that the manipulator mounting point is shifted from the center of the platform. It allows to choose the parking and operation configurations in such a way that the resulting CoM is the same for both (in the horizontal plane), see CoM position in Fig. \ref{fig:parking_fin}.

\subsubsection{Winches}
\label{subsubsec:winches}
\begin{figure}[t]
	\centering
	\includegraphics[width=1\linewidth]{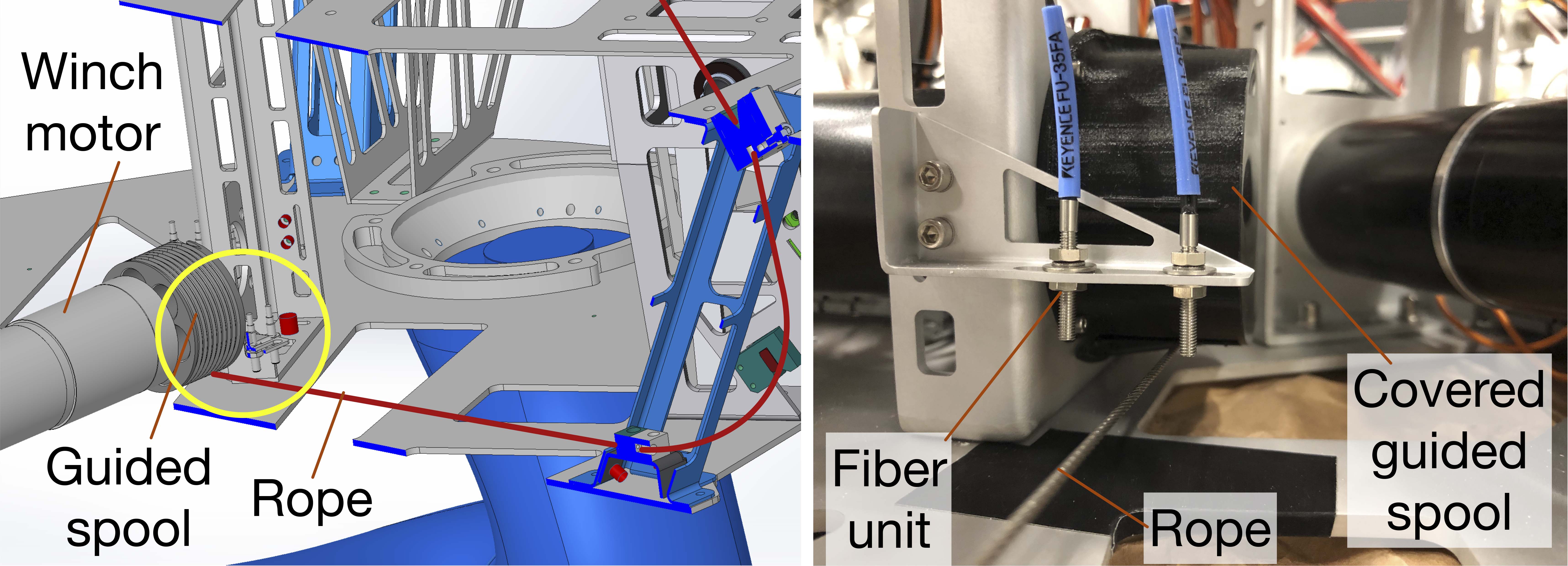}
	\caption{Winch structure: left subfigure shows the whole operation scheme, and right subfigure presents the detection system, which is inside of the yellow circle in left subfigure.}
	\label{fig:winches}
\end{figure}

a winch is a mechanical device that can control the length of the wire rope. Each winch consists of a DC-motor with installed guided spool and cover (Fig. \ref{fig:winches}). The wire rope is wound on the spool. Due to the presence of the brakes, no energy is required to keep the rope winding angle during operation work.

The \platformName ~contains three Maxon motor-based winches which help to maintain the CoM below the hanging point via controlling three DoFs, i.e., roll, pitch, and height of the platform. Control of the height using winches allows us to completely exclude aerial carrier from the aerial manipulation. 

Optical fiber sensing system from Keyence is used in order to remain within the operational point of the guided spool and to calibrate winches. By sensing light beam interruption and reflection, this system allows to detect an approaching rope to a reflection fiber unit (light source), see Fig. \ref{fig:winches} (right). In the calibrated configuration, wire rope is located in the middle between two light sources. The total length of the winch wire rope in the workspace of the spool is about 1.5 meter.

\subsubsection{Propulsion units}
\label{subsubsec:prop}

the cable-suspended platform is equipped with 8 propulsion units to resist disturbances induced by the robotic manipulator and to stabilize its own dynamics. From previous research \cite{ryll2016modeling, tognon2018omnidirectional}, it has been shown that by installing propulsion units in the special arrangement (not collinear), an omnidirectional 6 DoF wrench can be realized\footnote{
	It would be worthwhile to mention that, the minimum required number of propulsion units to realize 6 DoF wrench equals 7. We decided to use 8 units in order to address the exploitation of redundancy in the future studies. 
}. More specifically, the body wrench $\bm{w}$ and the thrust input vector $\bm{u}$ can be related by allocation matrix $\bm{A}$:
\begin{align}
\label{eq:allocation_map}
\bm{w} = \bm{A} \bm{u}.
\end{align}
\label{app:legs}

\begin{figure}[t]
	\centering
	\includegraphics[trim={0cm 0cm 9cm 0cm},clip, width=1\linewidth]{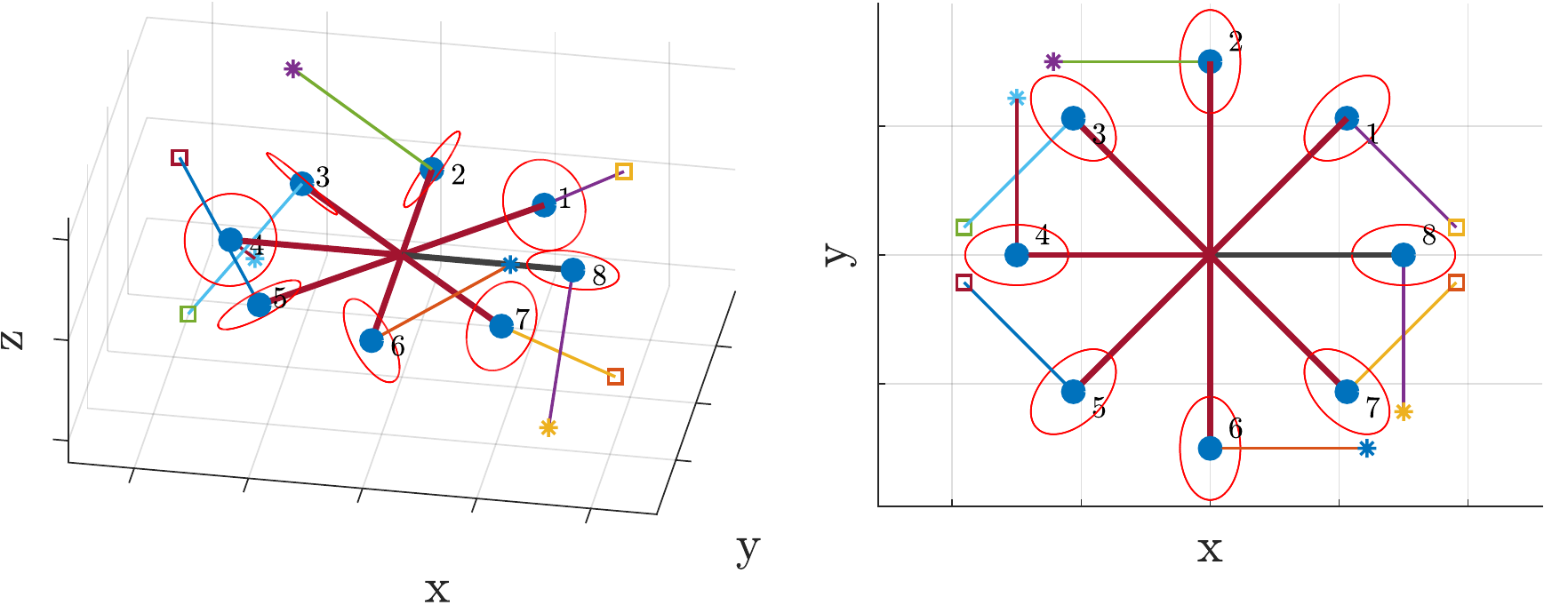}
	\caption{Optimized omnidirectional design for the propulsion units of the \platformName. The blue spheres represent the placement of the propulsion unit motors. The colored lines point the thrust direction of each motor. The star symbol denotes the counterclockwise propeller, and the square indicates clockwise propeller.}
	\label{fig:plotperspective}
\end{figure}

Following the line of~\cite{tognon2018omnidirectional}, to design the desirable propulsion unit arrangement, an optimization was carried out. In particular, the cost function to minimize was chosen as the condition number of the allocation matrix in order to guarantee the equal distribution between the propulsion units of the effort required to generate an omnidirectional wrench.
Additionally, the following constraints were considered: (i) the imposition of a particular (the unitary) eigenvector for the allocation matrix
 in order to obtain a balanced design; (ii) the normalization constraints for the unit vector defining the directions of the thrusters; (iii) an imposition of minimal installation angle of the propulsion units around the frame arm ($\alpha$) which guarantees the lifting of the landing legs with attainable motor thrust;
(iv) the perpendicularity between the thrust directions and the frame arm axes for an ease of mechanical implementation, i.e., $\beta$, angle around the perpendicular to the frame arm direction in the platform plane, is equal to 0. 
Mathematical treatment of (i) and (ii) can be found in \cite{tognon2018omnidirectional}, while those of (iii)-(iv) as well as the illustration of $\alpha$ and $\beta$ angles are given in the Appendix A. The result of optimization, w.r.t. the angle definition is: $\alpha_i = [53.1, -54.2, -126.9, 125.9, 53.1, -54.1, -126.9,  125.9]$ degrees and $\beta_i = 0$ degrees, where $i \in \{1,\dots,8\}$. The obtained design for the propulsion units is represented in Fig.~\ref{fig:plotperspective}.

\begin{figure}[b]
	\centering
	\includegraphics[width=1\linewidth]{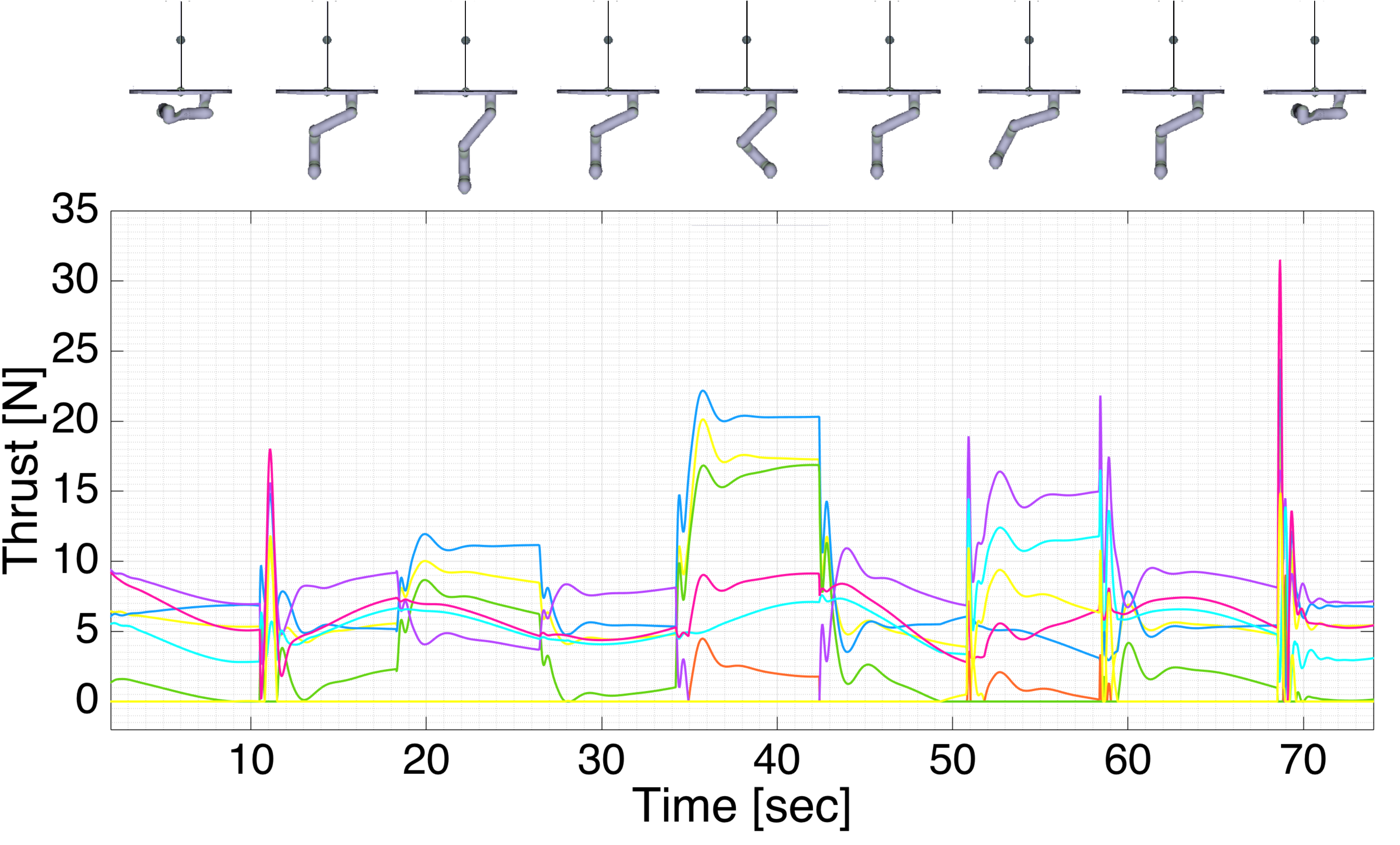}
	\caption{Simulated thrust values for 8 propulsion units required to compensate disturbances caused by the manipulator movement with high speed in joints.}
	\label{fig:thrust}
\end{figure}

The remaining design parameter, which is not considered in the optimization problem, is the required thrust value per motor. To calculate this value, desirable wrench $\bm{w}$ (and consequently the thrust value $\bm{u} = \bm{A}^\dagger \bm{w}$) was estimated in the simulation to compensate disturbances caused by the next sequence of manipulator postures: parking, operation, pick and place, operation, pick and place with different configuration of the robotic arm, operation, stretching, operation, parking, as shown in Fig.~\ref{fig:thrust}. As it can be seen, the maximum continuous thrust value required per motor is around 20\,$\mathrm{N}$. Selecting a maximum value of 40\,$\mathrm{N}$ for each motor guarantees some margin toward additional payload and disturbance. The contribution of the winches is neglected, hence the choice for the minimum thrust is conservative. To provide necessary thrust value, Kontronik Pyro 650-65 in pair with carbon 16x6'' propeller was chosen as the motor for the propulsion units.  In our experimental setup, this pair could exert continuous 50\,$\mathrm{N}$ thrust at 40\,\% throttle level.

\begin{figure}
	\centering
	\includegraphics[width=0.45\columnwidth]{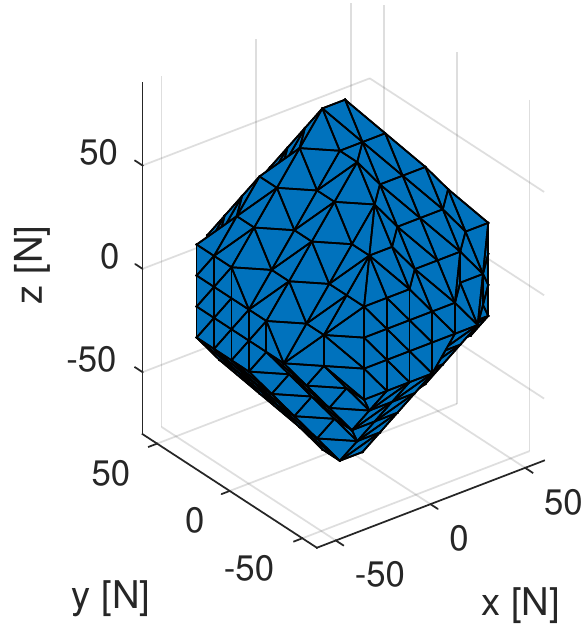}\
	\includegraphics[width=0.45\columnwidth]{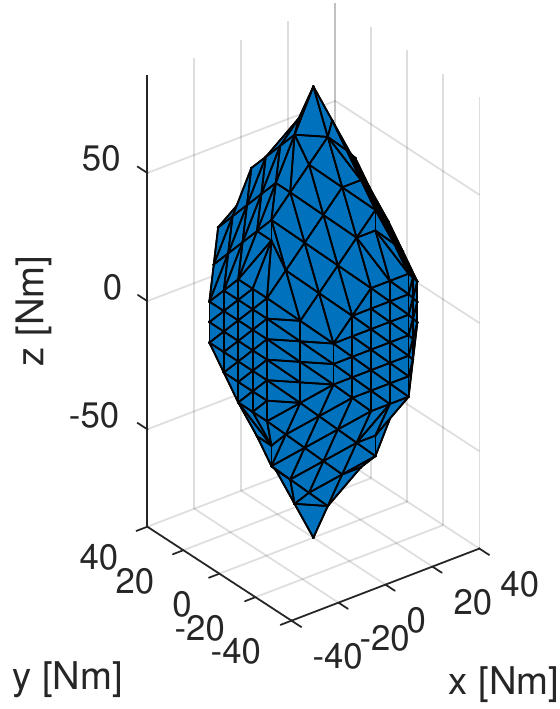}
	\caption{Set of admissible forces with zero torque (left) and set of admissible torques with zero force (right). }
	\label{fig:wrench}
\end{figure}

With the obtained propulsion unit configuration, the octo-rotor platform is capable to generate a set of independent forces and torques, thus allowing to decouple the control of the position and the orientation.
Fig.~\ref{fig:wrench} displays the set of body-frame admissible forces with zero torque (left) and the set of admissible torques with zero force (right). It is interesting to remark the high control authority around the yaw axis which is useful since this DoF cannot be exploited from the winches control.

\label{app:legs}

\

\label{app:legs}

\subsubsection{Mechanical design and dimensioning}
\label{subsubsec:mech}
a weight reduction is one  of  the main challenges in the design and production of aerial manipulators. Thus, the general mechanical frame including arms, plates, ribs, and motor holders is manufactured from insulated light-weight aluminum. Applying a set of methods for the weight optimization in the mechanical structure after several iterations, we managed to achieve the total weight of the \platformName ~with the installed manipulator, electronics, and batteries less than 45 kg.

As it can be seen in Fig. \ref{fig:main}, the \platformName ~consists of the two plates, and almost all components are installed inside the platform. Thus, winch motors and redundant robotic arm are mounted on the bottom plate, while other electronic blocks are either attached to the top plate or installed on the ribs between two plates. Possibility to remove the top plate provides easy and fast access to all electronic components. 
Propulsion unit motors are installed on the frame arms. The distance between the center of the platform and motor CoM is 0.75 meter. The orientation of each propulsion unit can be adjusted in the range of 0 to 360 degrees around $\alpha$ and +/-12 degrees around $\beta$, using a clamping mechanism.

\subsection{Electronics, sensors, and communication}
\label{subsec:arch}

\begin{figure}[t]
	\centering
	\includegraphics[width=1\linewidth]{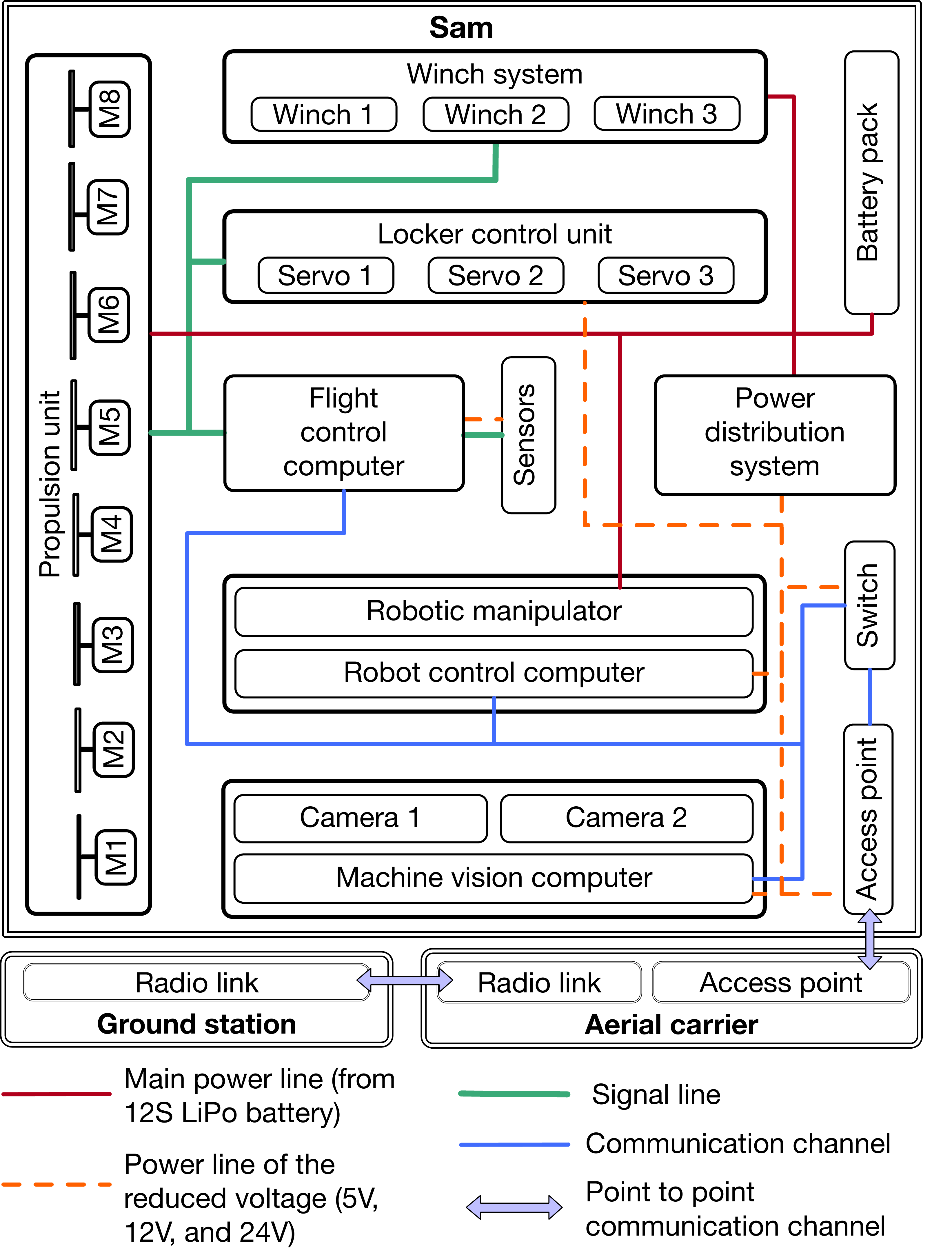} 
	\caption{The architecture of the \platformName.}
	\label{fig:arcitecture}
\end{figure}

The architecture of the \platformName ~is presented in Fig. \ref{fig:arcitecture}. As it can be seen, three different computers are installed onboard: robot control computer, machine vision computer, and flight control computer. The flight control computer is responsible for the control of the winches and propulsion units. Also, it is directly connected to the set of sensors for estimation of platform state 
during the operation: inertial measurement unit (IMU), GPS with real-time kinematics (commonly known as RTK), and two cameras for the machine vision system. It is worth noting that cameras have a dual-use. Their second purpose is to provide visual feedback to the human operator during teleoperation tasks. The connection between computers is established via a switch, which is connected to the access point. Through the access point, the point-to-point communication channel between the \platformName ~and the main aerial carrier is established. Additionally, there is a communication channel between the aerial carrier and a ground station via a radio link.

To power the whole system, 48 V with 100 Amps input is required for the worst case in terms of power consumption. As a battery pack, SLS 12S 21000 mAh is used. Thus, this battery pack can maintain the \platformName ~during time period of:
\begin{align}
 \frac{V \cdot Capacity}{P_{consumption}} = \frac{12 \cdot 3.7 \cdot 21 \cdot 60}{48 \cdot 100} = 11.65 \: [min]. 
\end{align}

\begin{figure}[]
	\centering
	\begin{subfigure}[]{0.45\linewidth}
		\centering
		\includegraphics[width=1\linewidth]{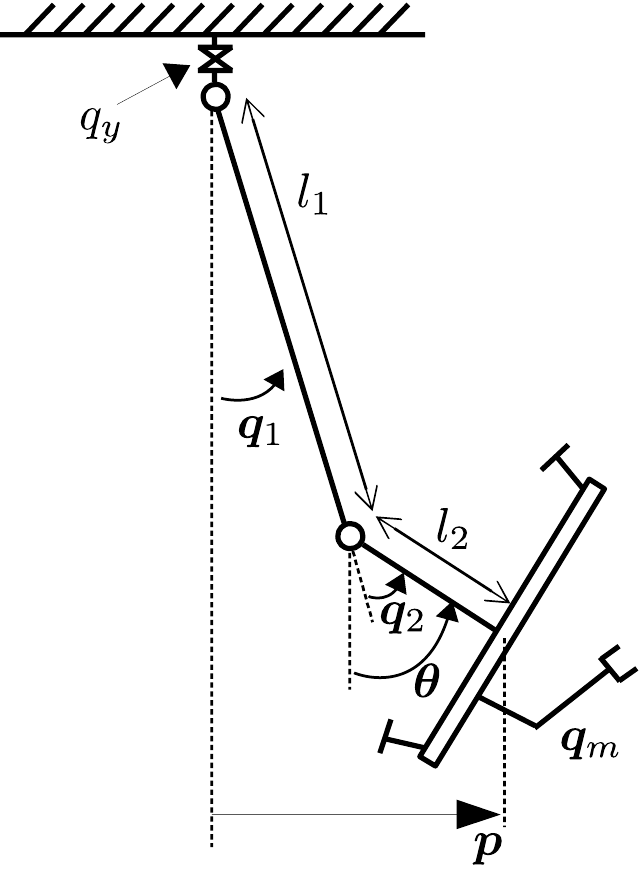}
		\caption{Coordinate configuration}
	\end{subfigure}
	\begin{subfigure}[]{0.51\linewidth}
		\centering
		\includegraphics[width=1\linewidth]{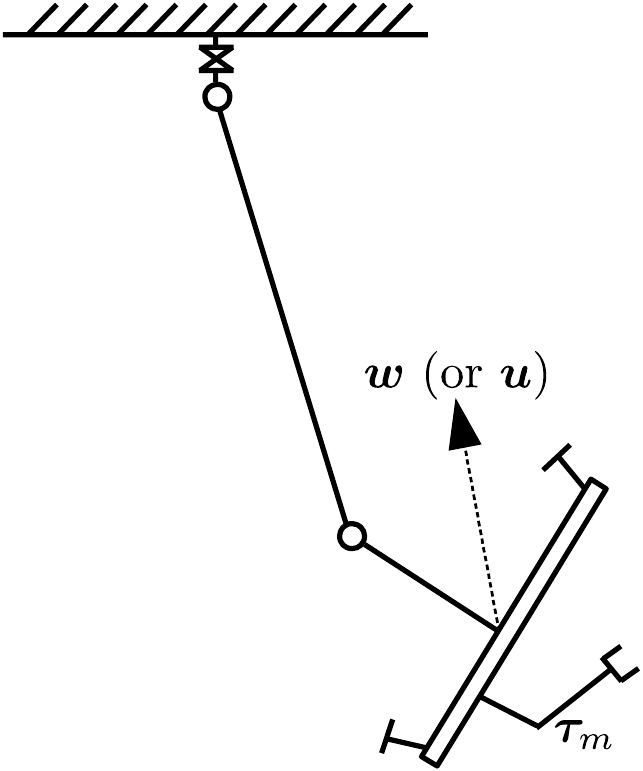}
		\caption{Actual control inputs}
	\end{subfigure}
	\caption{A schematic diagram of the \platformName.}
	\label{fig:sys_scheme}
\end{figure}

\section{Control Strategy}
\label{sec:control}

This section presents modeling, preliminary control scheme, and initial experimental results for the \platformName. Since the main aerial carrier is just hovering during the manipulation task (recall Section \ref{sec:Introduction}), it is assumed to be fixed. Carrying SAM to the desired position can be  interpreted as a slung-load transportation problem, which is already well studied in the literature, e.g., see \cite{bernard2009generic}. Also, winches are not considered in this section as they will be controlled by the powerful servomotors with a brake system to maintain the position. 

\subsection{Modeling of the \platformName}
\label{subsec:model}
In our modeling, the aerial carrier is neglected as it is hovering, cables are considered as massless rigid links with passive spherical joints, and the platform is approximated as a homogeneous disk.  Thus, the \platformName ~can be mathematically expressed as:
\begin{align}
\label{eq:dynamics_model}
\bM (\bq) \ddot{\bq} + \bC (\bq, \dot{\bq}) \dot{\bq} + \bg (\bq) = \btau,
\end{align}
where $\bM$ is the inertia matrix, $\bC$ is the centrifugal/Coriolis terms, and $\bg$ is the gravity vector. The configuration $\bq$ is:
\begin{align}
\bq =
[q_y \; \bq_1^T \; \bq_2^T \; \bq_m^T]^T.
\end{align}
As shown in Fig. \ref{fig:sys_scheme}a, $q_y \in \Re^1$ represents the total yaw angle, $\bq_1 \in \Re^2$ and $\bq_2 \in \Re^2$ represent the roll/pitch angles of the first and second passive joints, respectively. $\bq_m \in \Re^7$ represents the joint angles of the robotic manipulator. 

The control input $\btau$ can be written as:
\begin{align}
\btau =  [\tau_y \; \btau_1^T \; \btau_2^T \; \btau_m^T]^T.
\end{align}
Note that, while $\btau_m$ is the actual joint torque input of the manipulator, the others ($\tau_y$, $\btau_1$, and $\btau_2$) are virtual control actions in joint level, as there are no collocated actuations for the passive joints, see Fig. \ref{fig:sys_scheme}b.

Propulsion units can be used to generate these virtual actions by mapping them into the body wrench $\bm{w}$ using:
\begin{align}
	\bm{w} = \bm{J}^T  [\tau_y \; \btau_1^T \; \btau_2^T]^T,
\end{align}
where Jacobian $\bm{J}$ maps a body twist $\bm{v}$ into the roll, pitch, and yaw (RPY) rates, i.e., $[\dot{q}_y \; \dot{\bq}_1^T \; \dot{\bq}_2^T]^T
= \bm{J} \bm{v}$.
Finally, the desired body wrench $\bm{w}$ can be realized by eight propulsion units through the inverse of the allocation matrix (\ref{eq:allocation_map}).

\begin{figure}[]
	\centering
	\includegraphics[width=1\linewidth]{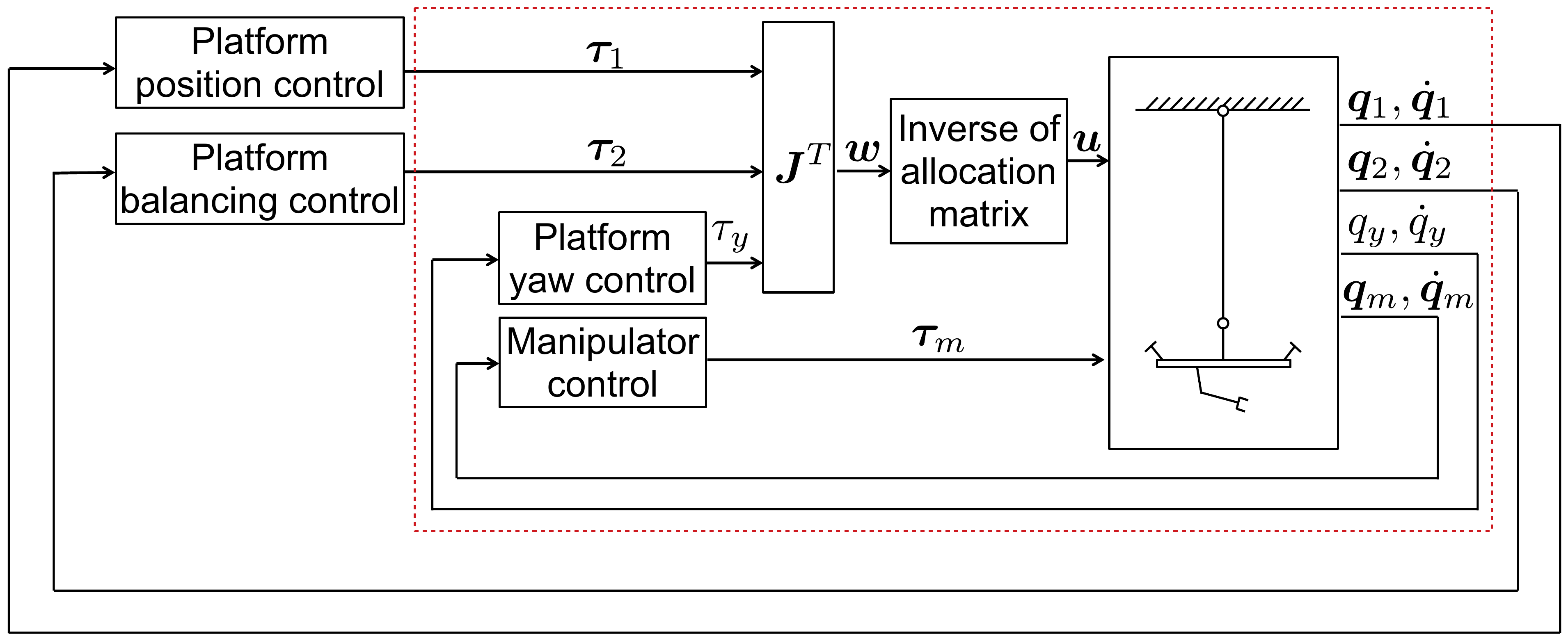}
	\caption{The cascade control scheme.}
	\label{fig:control_scheme}
\end{figure}

\subsection{Control design}
\label{subsec:controldesign}
Cascade control scheme in Fig. \ref{fig:control_scheme} is applied to the \platformName ~platform. Note that the manipulator has high sensing and control frequency, and that the yaw axis has the most significant control authority among torques (recall Fig. \ref{fig:wrench}). Therefore, the manipulator and yaw controls can be considered as inner loops of the control cascade, and both of them are performed using the proportional-derivative (PD) scheme with gravity compensation.

\subsection{Experimental results}
\label{subsec:valid}
This subsection presents the experimental results of the cable-suspended platform control around a vertical axis. To this end, two experiments for the yaw control of the  \platformName ~system using propulsion units were conducted: setpoint regulation with and without external perturbations. During both experiments, the robotic arm was not attached to the platform since its presence makes no difference for validation. Yaw measurements were obtained using IMU sensor.

\begin{figure}[t]
	\centering
	\includegraphics[width=1\linewidth]{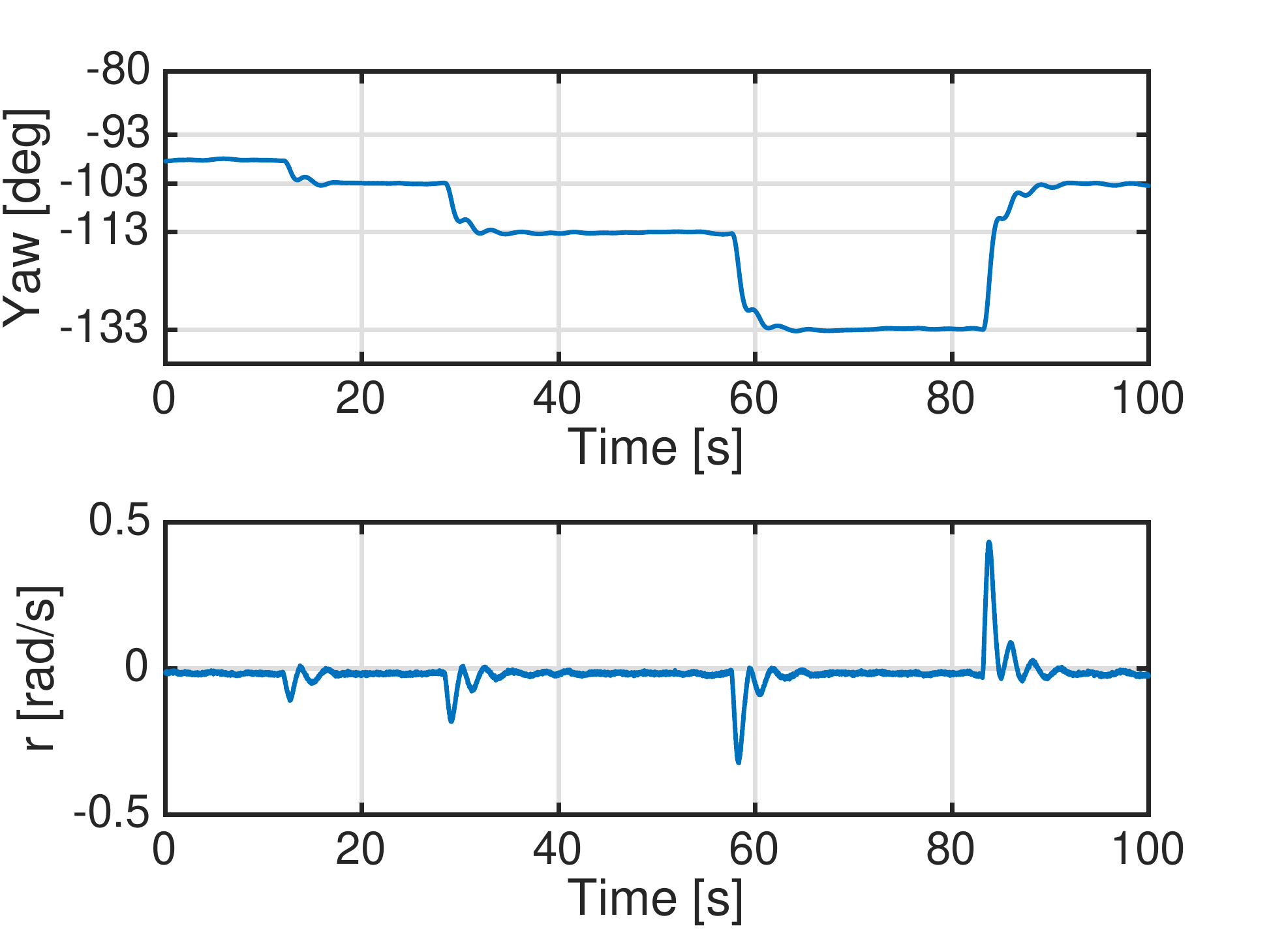}
	\caption{The response of the \platformName ~on the yaw commands using setpoint regulation control. }
	\label{fig:exp1}
\end{figure}

As it can be seen in Fig. \ref{fig:exp1}, during the first experiment the \platformName ~follows the desired set of yaw angles $q_y =[-98 ; -103 ; -113 ; -133 ; -103]$ degrees (upper plot). The PD gains were chosen to show a slightly over-damped behavior. The corresponding angular velocity $r$ of the platform around the vertical axis is shown in the same figure (lower plot). 

\begin{figure}[t]
	\centering
	\includegraphics[width=1\linewidth]{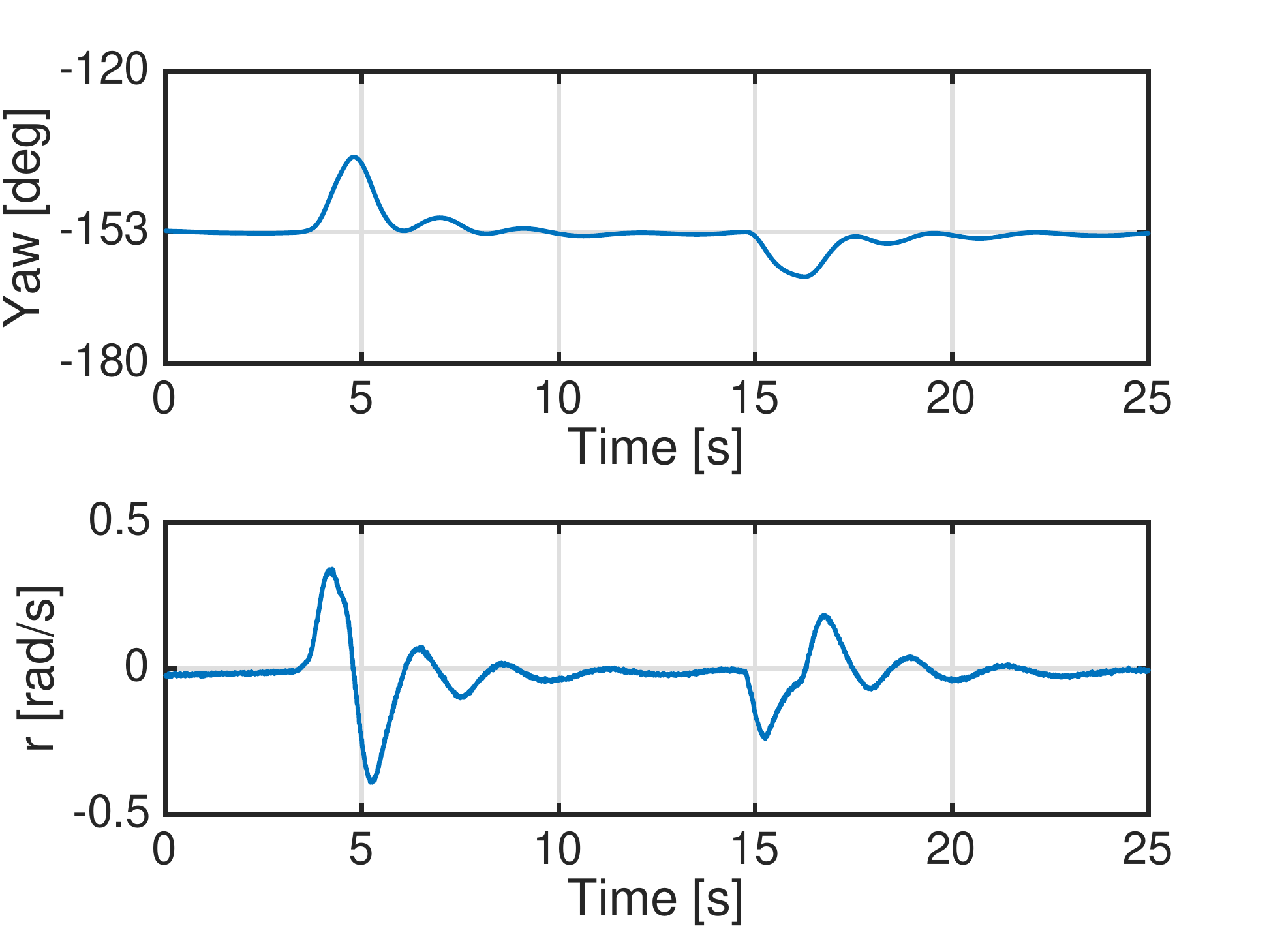}
	\caption{The reaction of the \platformName ~on the external disturbances around the vertical axis.}
	\label{fig:exp2}
\end{figure}

During the second experiment, the \platformName ~was externally disturbed by pulling the rope connected to one of the propeller arms, see Fig. \ref{fig:exp2}. 
Yaw angle of the platform converged to the desired value $q_y$\,$=$\,$-153$ degrees, and angular velocity $r$ converged to the zero despite external influence. These results validate that the \platformName ~can be controlled even with a simple control strategy. In the following link, a video clip which shows the aforementioned experiments as well as experiments with oscillation damping control (not shown in the paper) can be found: https://www.youtube.com/watch?v=ZoNNjQfUdJw.

\section{Conclusions}
\label{sec:end}
In this paper, a novel cable-suspended aerial manipulator SAM was presented. The \platformName ~has two actuation systems: propulsion units and winches. The main advantage of the developed platform w.r.t. existing solutions is the ability to perform aerial manipulation in a complex environment while keeping the main aerial carrier at a safe distance from the obstacles. Moreover, our solution is universal w. r. t. the carrier: UAV, manned aerial vehicle, or crane. Main aerial carrier supports the weight of the \platformName, and as a result, propulsion units of the platform can have reasonable dimensions since they should not compensate for the gravity.  Mechanical design, working principle, and architecture of all functional components were presented. Furthermore, a preliminary control strategy was proposed.



%
\section*{Appendix A}
\label{app:legs}

\begin{figure}[t]
	\centering
	\begin{subfigure}[t]{0.44\linewidth}
			\centering
			\includegraphics[width=1\linewidth]{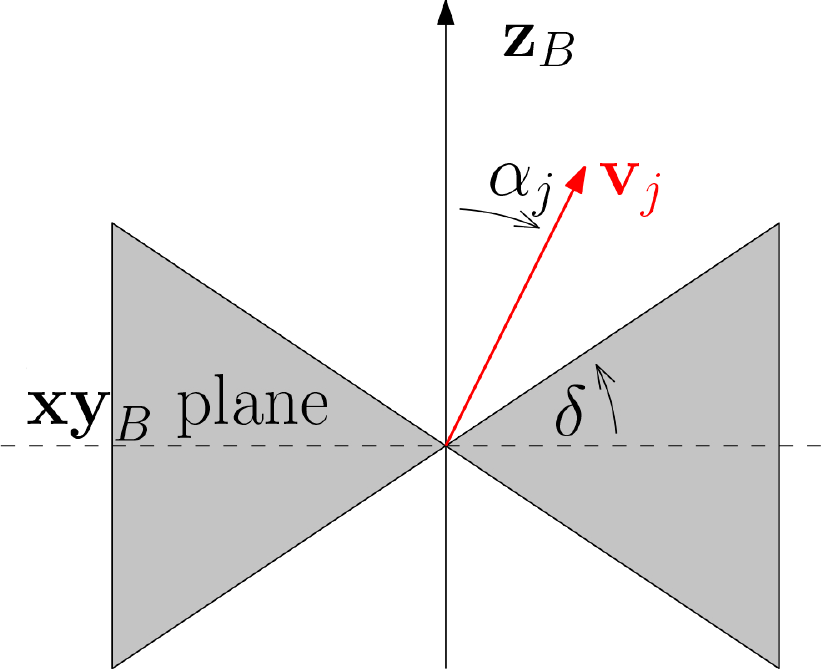}
			\caption{Illustration of the constraint (iii) in the optimization problem.}
			\label{fig:leg_constraints}
	\end{subfigure}
	\hfill
	\begin{subfigure}[t]{0.52\linewidth}
		\centering
		\includegraphics[width=1\linewidth]{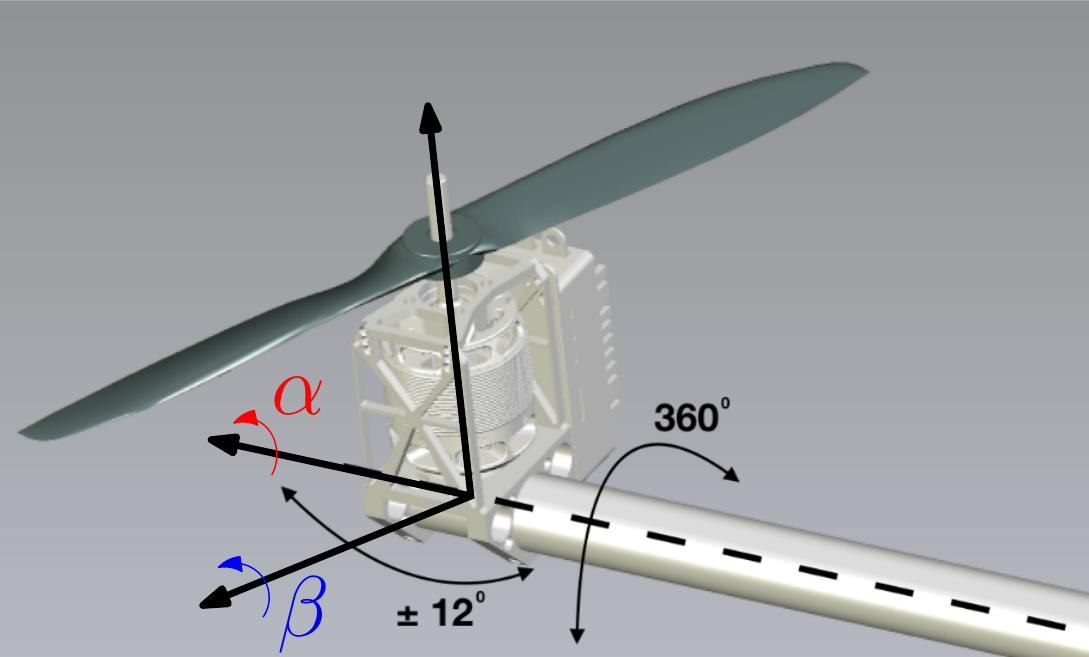}
		\caption{Definition of $\alpha$  and $\beta$ angles.}
		\label{fig:alpha_beta}
	\end{subfigure}
	\hfill
	\caption{Illustration of constraint (iii) (left) and installation angles (right).}
	\label{fig:appendix}
\end{figure}

The constraint (iii) is added to provide a minimum projection of the motor thrust vector along the z-axis of the body frame. 
 This constraint can be mathematically expressed as $|\mathbf{z}_B^T \mathbf{v}_j| \ge sin(\delta_p)$, where $\mathbf{v}_j$, $j\in \{2,5,8\}$ are the unit vectors indicating the direction of the propulsion unit thrusts expressed in body frame and related to the landing legs.  $\mathbf{z}_B$ is the z-axis of the body frame, and $\delta_p=\pi/6$ is the angle selected to allow the lifting of a leg of $1.5$\,$\mathrm{kg}$ with a leverage arm of $0.75$\,$\mathrm{m}$. Such a constraint prevents allocation of the thrust vectors in the gray area of Fig.~\ref{fig:leg_constraints} and is needed to ensure the lifting of the landing gear legs after the take-off. 

Constraint (iv) ensures $\beta_i = 0$ and can be written as $\mathbf{d}^T_i \mathbf{v}_k = 0$, $\forall i,k \in \{1,\dots,8\} $
where $\mathbf{d}_i$ are the vectors of the frame arms expressed in body frame. Such a constraint contributes to limit the mechanical complexity of the structure and also to avoid possible collision of the propellers with the frame arms. 

Convention of introduced angles $\alpha$ and $\beta$ is illustrated in Fig.~\ref{fig:alpha_beta}.

\section*{Acknowledgments}
We would like to thank Federico Usai (LAAS–CNRS/ Sapienza University) for his contribution in the design and simulation parts as well as Linus Grunert (Robo-Technology GmbH), Alexander Kreitmeyr (Robo-Technology GmbH), and Khizer Shaikh (Elektra Solar GmbH) for their contribution in design and manufacturing parts. 

We also thank our colleagues in the DLR Institute of Robotics and Mechatronics, especially those in electrical and mechanical workshops for their great support and maintenance.

\newpage

\bibliographystyle{ieeetran}
\bibliography{mybib}

\end{document}